# Improving PSO Global Method for Feature Selection According to Iterations Global Search and Chaotic Theory

**Shahin Pourbahrami**[1*]

Computer Engineering Department, Faculty of Electrical and Computer Engineering, University of Tabriz, Tabriz, Iran[1]

**Abstract**

Making a simple model by choosing a limited number of features with the purpose of reducing the computational complexity of the algorithms involved in classification is one of the main issues in machine learning and data mining. The aim of Feature Selection (FS) is to reduce the number of redundant and irrelevant features and improve the accuracy of classification in a data set. We propose an efficient ISPSO-GLOBAL (Improved Seeding Particle Swarm Optimization GLOBAL) method which investigates the specified iterations to produce prominent features and store them in storage list. The goal is to find informative features based on its iteration frequency with favorable fitness for the next generation and high exploration. Our method exploits of a new initialization strategy in PSO which improves space search and utilizes chaos theory to enhance the population initialization, then we offer a new formula to determine the features size used in proposed method. Our experiments with real-world data sets show that the performance of the ISPSO-GLOBAL is superior comparing with state-of-the-art methods in most of the data sets.

## 1. Introduction

In recent researches, the classification methods based on features selection have been studied to increase the classification performance. Currently, finding a set of informative features with small size and high accuracy is one of the drastic ways to solve the classification problem because it can reduce the training time of a learning task, brief the design of the classifier, and improve the classification accuracy. Thereby, feature selection process is applied to machine learning methods to select a high-quality feature set by eliminating the irrelevant and redundant features. Hence, different types of evolutionary algorithms such as Genetic Algorithms (GA) [1], Differential Evolution algorithms (DE) [2], Particle Swarm Optimization algorithms (PSO) [3, 11], Shuffled Frog Leaping Algorithms (SFLA) [4] and Evolutionary Computation (EC) techniques have been successfully applied for the feature selection. The feature selection task is a challenging issue due mainly to achieving reduced dimensions in the large search space [9]. The subsets with suitable features are a group of features that include an appropriate composition of ideal features that leads in high accuracy and ideal classification in performance. Guyon and Gheyas, show that there could be multi-way interactions among features [8, 9]. Regarding this point of view, feature selection methods are the bases for data mining to keep good features for improving learning tasks that produce higher classification accuracy alongside the ignoring of the most irrelevant features and less important ones [10]. Therefore, a universal search is practically impossible in most situations. Although many different search techniques have been used to feature selection, most of these techniques still suffer from finding an entirely global search technique [9, 11].

---

[*] Corresponding Author.
*Email Addresses:* sh.pourbahrami@tabrizu.ac.ir (Shahin Pourbahrami)

Particle swarm optimization is a relatively recent evolutionary algorithm, which is computationally more efficient than the other metaheuristic methods. Meanwhile, one of the most important significant techniques in the process of feature selection is PSO [5, 6, and 11]. But, there are still some major issues regarding the traditional PSO while is applied for the feature selection. The most important problem involves the question that: How far is the optimal solution from the primary population in random initialization? In fact, the problem lies in the fact that if the optimal answer is too distant from the predicted guess, it may not be feasible to reach the general optimal solution in the assigned time. Which in this respect Gutierrez et al. [7, 43] showed that random initialization strategies in PSO have been applied in different problems with high-dimensional search spaces. The second problem is that traditional personal best and global best updating mechanisms of PSO may results in missing some informative features. Therefore, the capability of PSO for feature selection has not been fully investigated and we will propose our method to further consideration in the initialization and the updating mechanisms in PSO that particles try not to converge to a single point. In order to better address global search problems, an efficient the ISPSO-GLOBAL method proposed. The ISPSO-GLOBAL enhances the search capability for finding an optimal solution. With this respect, our main contributions are specially summarized as follows:

1. We first introduce a new formula that employs a function to automatically identify the number of selected features in a bounded area.

2. The proposed method investigates a novel initialization strategy which utilizes chaos theory to enhance the population initialization particle swarm optimization.

3. Finding a good solution through an iterative process by using specific add and delete operations which were saved during special iterations in order to be used in next iterations (specially the last ideal iterations).

The reminder of the paper is organized as follows. Section 2 provides an overview of related works on feature selection. In Section 3, the ISPSO-GLOBAL method which is used for feature selection in this paper is discussed. In Section 4, the described ISPSO-GLOBAL method is applied to several real world data sets and the results are presented and "Experiments" provides the experimental results and discussion. Finally, Section 5 explains Conclusions of the paper.

## 2. Related Works

The aim of feature selection (FS) is choosing a limited number of features by which the redundant and unrelated characteristics with no change in performance level. In the presence of many redundant and irrelevant features, learning models tend to overfit and accordingly leads to reduce the classification accuracy. Four categories have been identified for feature selection algorithms including filter approaches [10], wrapper approaches [11], embedded approaches, and hybrid approaches [12]. Since filter approaches do not depend on any learning algorithm and make use of feature set statistical analysis, they are usually very fast. However, wrapper approaches aim at choosing the features that improve learning algorithm accuracy. Since wrapper approaches, a single learning model is utilized repeatedly for the purpose of evaluating different set of features, they tend to have a slower pace than filter approaches [8, 12]. As the name of hybrid approaches indicates, they utilize both filter approaches and wrapper approaches to identify and utilize the strengths of both approaches [10].

Filter approaches use two methods, namely univariate and mulrivariate methods. In univariate method, the features are classified according to a statistically-based criterion. However, multivariate methods classify the features according to two criteria of relevance and redundancy. Some univariate filter methods have been suggested: Information gain (IG) [13], Gain Ratio [14], Term Variance (TV) [15], Gini Index (GI) [16], Laplacian Score (L-Score) [17] and Fisher Score (F-Score) [18]. On the contrary, multivariate approaches use feature dependencies to evaluate feature relevance. For more illustration, in this category, Minimal-redundancy-maximal-relevance (MRMR) [19] is a well-known multivariate feature selection method that uses different strategies to generate subsets and progress the search processes. Regarding the feature selection, we can also mention the other methods called sequential forward search (SFS) and sequential backward search (SBS). SFS starts with a single feature and iteratively add & remove features until some terminating criterion is met Ref [20, 22]. While SBS starts its search process with full feature set and then goes on to do add & delete operations based on learning algorithms accuracy. Moreover, the sequential backward search attempts to find solutions ranged between suboptimal and near optimal regions so that they involve searching locally rather than globally [22]. On the other hand, finding the solution of optimal or near optimal is quite stickler because of search algorithms involving a partial search in the solution space. Therefore, the recent trend of research has been paid attention to meta-heuristics algorithms such as genetic algorithm (GA) [21, 22], particle swarm optimization algorithm (PSO) [23] [11, 25-28]. and ant colony optimization algorithm (ACO) [24].

Although the PSO has been shown as an effective approach for searching optimal final feature subsets, it suffers from several shortcomings. One of the problems with PSO-based feature selection methods is that they relinquish the number of features in their search processes while they only emphasize maximizing the classification accuracy. Another limitation is to select similar features in the final feature subset. Moreover, the existing PSO-based methods do not use correlation information of the features to guide the search process probability to be selected in the good features in during iteration, which increase the classifier performance. We propose a robust method to overcome the mentioned problems which considers generative models of search features structure. The method also incorporates informative features and iteration based methodology to infer good features and employs a statistical method to select the most appropriate model. Comparing with other methods, our method selects the smallest subsets with high classification accuracy during iterations process while searching global space solutions.

## 3. Proposed Method

In this section, our proposed method is presented in six steps. Step 1 determining the number of features ,step 2 Population Initialization, and Step 3 is about updating the particle positions suggested in [43], step 4 cover ideological principles of the present method, step 5 is about Gbest mutation suggested in [2, 30-34], and step 6 is about k-Nearest Neighbor (kNN) suggested in [45]. The purpose of the present study is answering the following research questions: In PSO-based feature selection, in what way the disadvantages of fine-tuning near local optimal points and early convergence points can be effectively removed? Fig.1 demonstrates the ISPSO-GLOBAL steps. In Fig.1 rectangles with dash lines show the steps proposed by ourselves and the others exploit from existed methods.

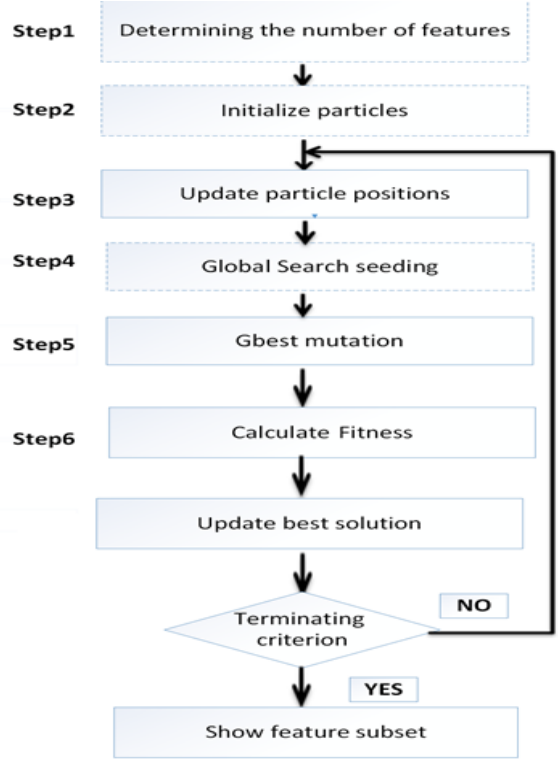

Fig.1. Flowchart of the proposed ISPSO-GLOBAL method.

## 3.1. Step 1- Determining the number of features

The purpose of this part is identifying the frequency of selected features in a limited zone:

$$K = \frac{(N-(vN))+CR}{CR-r} \qquad (1)$$

Where *N* denotes the number of features in a given data set. $r \in [1, 50]$ and $v \in [0, 1]$ are adjustable defined parameters The CR is the number of data samples. In this step, for every given data set and according to all samples and their features, the final features in the last phase of the algorithm will be obtained. Parameter r and v are adjustable by the user, and r can be determined according to the total number of samples in each data set. In large data sets, this parameter is considered to be nearly 50, and in small data sets the value is almost zero. Parameter v is the criterion for adjusting the percentage of the features favored by the user in the final output of the algorithm. According to different experiences with different data sets, in identifying the frequency of the features in the final output, this method has proved to bear better results than.

## 3.2. Step 2- Population Initialization

The most important stage in algorithm optimization is initializing the population since this step influences the speed of convergence and the subset of the final features. In this line, first some initial solutions which have already been produced start the meta-heuristic algorithms with the purpose of helping the population to move toward the optimal final solution(s). After some previously specified criteria are fulfilled, searching stops. The distance between the optimally obtained solution and the initial population determine convergence rate and the time needed for computation. The initial population is obtained based on the lost information and making random guesses. The worst state takes place when the random guess and the optimal answer completely mismatch each other, and the global best solution may not be obtained within the limits of the allotted time. Therefore, it is suggested to start with an initial population containing the best possible guesses (random guesses based on chaos theory). Chaos theory has the advantage of uniform distribution of the initial solutions enhancing the likelihood of convergence in the proposed method. A very recent version of ISPSO-GLOBAL is used in this part with the purpose of differentiating search moves if the best particle can hardly be found. We propose a novel chaos -based initialization strategy which improves space search and utilize chaos theory to enhance the population initialization. Chaos refers to a random state in the deterministic system. This state is the valuation of nonlinear systems through deterministic rules and could be a long-term behavior while not a fixed period. Chaos is susceptive to initial conditions. Chaotic convulsion will traverse the state's entire correspondent itself without repeating these states within a certain range. The use of chaos search is more advantageous than that of absolute random searches. Logistic and Tent maps are frequently used in chaos search [29, 41, 42]. The equation of the Logistic map is as follows:

$$X_{(t+1)} = \alpha X_{(t)} \times (1 - X_{(t)}) \tag{2}$$

Where $t$ is the iteration times; $X_{(t)} \in [0, 1]$; and $\alpha$ is the control parameter. If $\alpha=4$, then the system is in a state of chaos. The Logistic map belongs to Li-Yorker chaos [6].

$$X(t+1) = \begin{cases} 1 & if \ 0.5 \leq X_t \leq 1 \\ 0 & if \ 0 \leq X_t \leq 0.5 \end{cases} \tag{3}$$

### 3.3. Step 3- Updating the particle positions

PSO is a swarm intelligence technique proposed by Kennedy & Eberhart in 1995 [37, 38]. PSO simulates the social behavior such as birds flocking and fish schooling. In PSO, candidate solutions are represented as particles in the search space. Particles are moved through the solution space to looking for the optimal solution by changing the position of each particle based on its own experience and also using the position of its neighboring particles. The PSO was introduced for the optimization of problems in continuous multidimensional search space but PSO for feature selection are discrete. During the movement, the current position of a particle $i$ is represented by a vector $x_i = (x_{i1}, x_{i2},...,x_i)$, where $d$ is the dimensionality of the search space. The velocity of a particle $i$ is represented by $v_i = (v_{i1}, v_{i2}, ..., v_i)$, while is $v_i$ the velocity of the particle in $x$-th direction. It is worth mentioning that these velocities are limited by a predefined maximum velocity, $v_{max}$ and $v_{id} \in [-v_{max}, v_{max}]$. The best position has been seen previously by a particle is recorded as the Personal best known as *Pbest* and the best position that has been explored by the whole swarm so far is the Global best called *Gbest* [43]. In fact, PSO is seeking for the optimal solution through updating the position and the velocity of each particle according to the following equations:

$$\hat{v}_i(t+1) = v_i(t) + c_1 \times r_{1,i} \times rand(p_i(t) - x_i(t)) \times v_i(t) + c_2 \times r_{2,i} \times rand(p_{gd}(t) - x_i(t)) \tag{4}$$

$$s(v_i) = sigmoid(v_i) = \frac{1}{1+e^{-v_i}} \tag{5}$$

$$\text{if} \quad \text{rand} < s(v_i(t+1)) \quad \text{then} \quad x_i(t+1)=1 \quad \text{else} \quad x_i(t+1)=0 \tag{6}$$

$c_1$ and $c_2$, were set to be 1.5 and 2 respectively. Moreover, the upper and lower bounds for $v$ were set to 4 and -4, respectively.

### 3.4. Step 4- Global Search seeding

In this step, we make inquiries regarding the performance of the subset with an objective function for choosing the most distinct and informative features considering measures the between- features correlation of the features [39, 40]. The goal of finding the relationships between features in this step is to exploit salient features and save them in storage list (based on number repeat features). Later, after choosing P feature from Boltzmann Distribution Transport Equation, they can be divided among the new particles which have been generated.

In statistical mechanics, the Boltzmann distribution could be probability distribution that provides the probability a system which will be in a certitude state as a function of that state's features and the temperature of the system. It is given as:

$$P(vots_{id}) = \frac{e^{-vots/T}}{\sum_{j=1}^{M} e^{-vots/T}} \tag{7}$$

We postulate the probability of a given state as a function of the voting features; it must satisfy the relationship mentioned above for all possible values of $vots_{id}$. Therefore, we will investigate specified iterations during our work method for production salient features and store them in a storage list. Note that the goal is to find informative features based on frequency in specified iterations features determination occurs with top fitness.

To overcome the global search and diversity issues in PSO-based methods, the proposed method determines fixed number features and applies the Boltzmann distribution by maintaining the diversity of the population in steps of primeval and conduction particles to goodness solutions and storage least correlation features in the list of storage to be used in next generation. In addition, the proposed method finds good solutions in its last iterations by employing add and delete operations during its iterations. Our method counts numbers of repeated informative features with high fitness in several iterations and keep it in a storage list and then they are added or deleted by using the Boltzmann distribution. Finally, these features are applied inside binary vectors to be optimized though an optimization process.

Thus, the degree and direction of linear relationship among two variables are returned by Pearson correlation. The stronger the linear relationship between the two variables, the greater the correspondence between changes in the two variables. When there is no linear relationship, there is no covariability between the two variables, so a change in one variable is not associated with a predictable change in the other variable [44]. In this study, the Pearson correlation coefficient is used to measure the correlation between different features as follows:

$$c_{ij} = \frac{\sum (x_i - \bar{x}_i)(x_j - \bar{x}_j)}{\sqrt{(x_i - \bar{x}_i)^2} \sqrt{(x_j - \bar{x}_j)^2}} \tag{8}$$

$$cor_i = \frac{\sum_{j=1}^{f}|c_{ij}|}{f-1} \quad if \quad i \neq j \tag{9}$$

In the example shown in Fig. 2, for five iterations the number of informative features having high fitness are sorted and saved in the storage list (based on number frequency features in these iterations). In the following step, when we intend to choose informative features, we select them from the storage list using Boltzman formula and move them to the new generation by add operation. In new generation, we can delete the low-fitness features using delete operation.

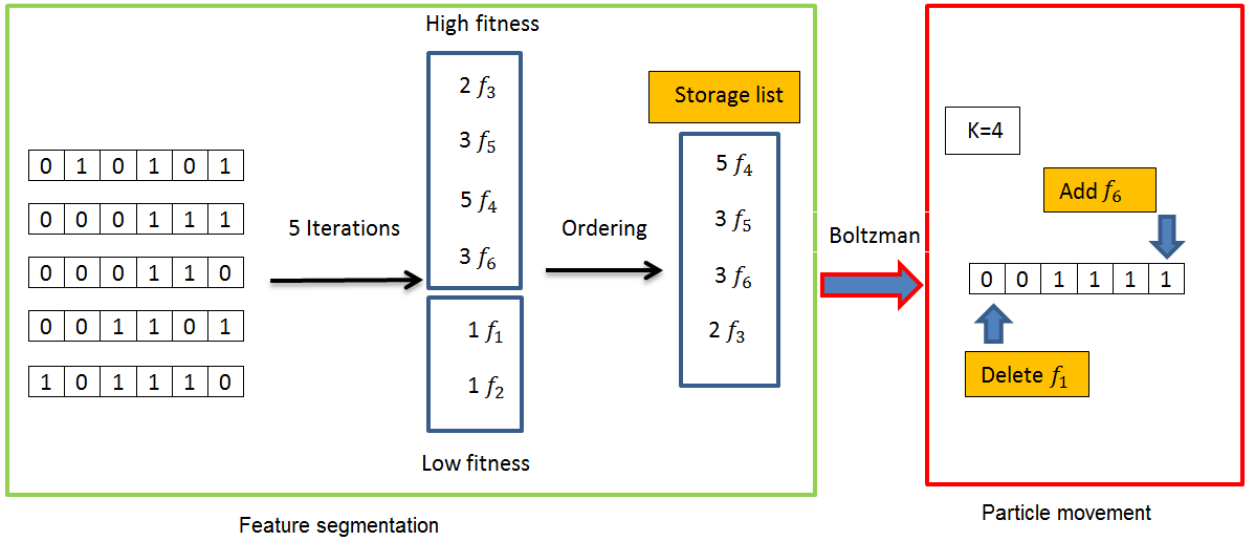

Fig. 2. Illustration of the global search strategy of the proposed method.

### 3.5. Step 5- Gbest mutation

The most recent position of particles is mainly influenced by the best solution it has (Pbest) as well as the best solution of the swarm labeled as Gbest. However, if Gbest position is not distant from suboptimal solution, and the swarm moves to suboptimal position, the mentioned type of movement would be deficient and very subversive. The researchers have suggested [2, 30-34] causing small mistakes randomly or consider the particles moving in the reverse direction for the purpose of identifying the regions that have not been explored and relieving the suboptimal position. For the purpose of creating diversity, in this study mutation or reversing Gbest particle is utilized. In other words, in this study Gworst or the worst fit particle is replaced by the muted particle or reverse Gbest. In this way, the unexplored regions are discovered by the use of Gworst particle. Meanwhile, Gbest particle reserves global optimal solution since there is no change in Gbest particle, by itself. In this line, the researchers used a parameter and applied mutation only if the parameter value is less than 1. The input for the mutation process included Gbest and Gworst. Gbest particle is kept in a variable which is not permanent and in order to replace Gworst, it is mutated. In order to ensure mutation, all dimensions of the impermanent variable are examined. In other words, in case the mutation in any of the dimensions is less than mutation probability, for the purpose of ensuring mutation of the selected dimension, a random number is given. If mutation rate was supportive of the quantity of the dimensions aimed at mutation, that rate should be selected immediately so that some extent of variation will be introduced in the particle instead of introducing excessive randomization within it. Lee et al. showed that this method is effective in BPSO [35].The mutation probability of this method is 1/d indicating that in the binary vector at least one of the bits has been changed [36].

## 3.6. Step 6- Fitness Function

For the purpose of evaluating each of the solutions, KNN (K-nearest neighbor classifier) is used in this study [45]. Before the evaluation process, first of all, each feature is normalized being scaled between -1 and 1. The normalization method changes the dominating features with bigger numeric values to those with finite numeric ranges. Experimental results incontestable that scaling the feature values will facilitate improve the classification accuracy. Thus, in our method, a linear normalization method was applied to scale the dataset as follows [47]:

$$x = lower + \left[(upper - lower) * \left(\frac{value - value_{min}}{value_{max} - value_{min}}\right)\right] \quad (10)$$

In this formula, the lower and upper bounds of the process of normalization are indicated by *upper* and *lower*. The highest and lowest values for feature *x* are also indicated by $value_{max}$ and $value_{min}$. After the process of normalization finishes, a new data set is obtained from the original data set for each candidate solution. Later by the use of KNN method performance is evaluated by the use of 10-fold cross validation method [46]. Therefore, with the aim of obtaining a proper evaluation of model prediction performance (or estimating the fitness value of the article) initially the first nine data sets were used in the process of training; however, the last data set was used in the process of validation. If the two obtained solutions revealed almost the same performance, the solution with the least number of features would be chosen [47].

## 4. Experiments and results

The effectiveness of considering ISPSO-GLOBAL method is tested using twenty well-known and frequently used datasets with varying sizes that are retrieved from UCI repository. The results from this study and those of the prevalent and famous methods are compared for metrics including Precision (P), Recall (R), and F-measure (F). The Precision, measures total variety of correct positive exceptions to the entire varieties of positive exceptions and Recall, measures total variety of correct positive exceptions to the entire number of positive documents. However, F-score could be a harmonic combination of P and R. These evaluation metrics are defined in Eqs. (11) and (12):

$$P = \sum_{i=1}^{n} \frac{TP_i}{TP_i + FP_i} \quad (11)$$

$$R = \sum_{i=1}^{n} \frac{TP_i}{TP_i + FN_i} \quad (12)$$

Where FP, FN, TP, and TN refer to candidate false positives, false negatives, true positives, and true negatives respectively. Finally, F-measure is defined as follows:

$$F = \frac{2 \times P \times R}{P + R} \quad (13)$$

In order to illustrate the capability of the proposed method, kind of standard measurements, namely accuracy, times, and F-measure, were applied [48-50].

## 4.1. Experiments to Analyze the Execution Time

According to the results obtained from PSO, CBPSO [51], HPSO-STS [52], SPSO-QR [53], PSO (4-2) [54] and HPSO-LS [56] algorithms as well as the results of different experiments indicate that ISPSO-GLOBAL method is a reliable and valid method for feature selection. Meanwhile, the time needed for conducting the present method and wrapper methods such as CBPSO, PSO-STS, HPSO-STS, SPSO-QR, and PSO (4-2) are compared with each other. Furthermore, the time needed for executing ISPSO-GLOBAL is also compared with the time needed for filter methods such as IG, F-Score, TV, and mRMR. MATLAB software is used to code the algorithms. The software was installed on a computer with an Intel Core 4 Due, 3 GHz CPU and 4 GHZ RAM.

## 4.2. Datasets

Twenty datasets are adopted in our simulation experiments in order to validate the performance of ISPSO-GLOBAL in comparison with methods. These datasets are taken from the UCI machine learning repository [55], including; "Iris", "Thyroid", "Liver", "Pima", "Glass", "Vowel", "Wisconsin Breast Cancer (WBC) ", "Wine", "Heart", "Segment", "Two norm", "Sonar", "Lung cancer", "LSVT Voice Rehabilitation", "Arrhythmia", "Protein", "Zoo", "Vehicle", "Lymphoma", "Wdbc" and "Soybean". Numerous instances of small, medium, and large data sets in the mentioned machine have been used in various studies on machine learning. A descriptive summary of these datasets is presented below and a tabulated summary is presented in Table 1.

**Table 1**
Description of used Data sets.

| name | #features | #patterns | #classes |
|---|---|---|---|
| **Iris** | 4 | 150 | 3 |
| **Thyroid** | 5 | 215 | 3 |
| **Liver** | 6 | 345 | 2 |
| **Pima** | 8 | 768 | 2 |
| **Glass** | 9 | 214 | 6 |
| **Vowel** | 10 | 528 | 11 |
| **Wisconsin Breast Cancer (WBC)** | 10 | 699 | 2 |
| **Wine** | 13 | 175 | 3 |
| **Heart** | 13 | 270 | 2 |
| **Segment** | 19 | 2310 | 7 |
| **Sonar** | 60 | 208 | 2 |
| **Two norm** | 20 | 7400 | 2 |
| **Vehicle** | 19 | 846 | 4 |
| **Protein** | 20 | 116 | 6 |
| **Zoo** | 16 | 101 | 7 |
| **Soybean** | 35 | 47 | 4 |
| **Wdbc** | 30 | 569 | 2 |
| **Lung cancer** | 56 | 32 | 3 |
| **LSVT Voice Rehabilitation** | 309 | 126 | 2 |
| **Arrhythmia** | 279 | 452 | 16 |
| **Lymphoma** | 4026 | 47 | 2 |

### 4.3. Evaluation mechanism

Several experiments were performed to evaluate the performance of the proposed ISPSO-GLOBAL method. Fig.2, compares the accuracy of the proposed method with well-known wrapper methods include; GA, SA, ACO and PSO methods. The obtained results indicated that the method adopted in this study was able to obtain greater accuracy than all other methods. As an instance, in the data set labeled as Vowel, the recorded run time for ISPSO-GLOBAL was 3.41 s. However, for SA, GA, ACO, and PSO, the obtained run times were 7.58, 4.13, 90.80, and 1.71 s, respectively. Moreover, the features chosen by each individual algorithm during 20 different runs is shown by Fea.NO. Form the results it can be seen that in most cases the proposed method require lower computational resources compared to the others.

**Table 2**
Average of running time (in seconds). In each data set, the best method is marked by the boldface.

| data sets | GA | SA | ACO | PSO | ISPSO-GLOBAL |
|---|---|---|---|---|---|
| Iris | 0.23 | 0.42 | 0.56 | 0.21 | 0.45 |
| Thyroid | 0.54 | 0.46 | 1.71 | 0.31 | **0.27** |
| Liver | 1.29 | 1.01 | 4.88 | 0.56 | **0.52** |
| Pima | 7.44 | 5.77 | 60.27 | 3.15 | 2.53 |
| Glass | 0.59 | 0.46 | 9.46 | 0.34 | 1.06 |
| Vowel | 4.13 | 7.58 | 89.51 | 1.71 | 3.41 |
| WBC | 7.19 | 13.08 | 90.80 | 3.05 | 2.41 |
| Wine | 0.67 | 1.09 | 18.90 | 0.62 | 0.66 |
| Heart | 1.46 | 0.90 | 21.06 | 0.57 | 0.87 |
| Segment | 2:12:19 | 1:43.67 | 88.50 | 56.62 | **0.63** |
| Sonar | 2.68 | 5.08 | 56:17.53 | 1.09 | 20.62 |
| Two norm | 20:57.33 | 28:31.80 | 35:05:06.14 | 23:00.75 | **5:13.54** |

**Table 3**
Compare subset feature selected by algorithms and show Fea.NO in algorithms.

| data sets | GA | SA | PSO | ISPSO-GLOBAL |
|---|---|---|---|---|
| Iris | 1,2,3 | 1,2,3,4 | 2,3 | 3,4 |
| Thyroid | 1,2,3,4 | 2,3,4 | 2,3,4,5 | **4,5** |
| Liver | 2,3,4,5 | 1,2,3,5 | 2,3 | **2** |
| Pima | 1,2,3,4,5,7 | 1,2,3,4,5 | 4,5,6 | **3,9** |
| Glass | 2,3,4,5,6 | 3,4,5,6 | 4,5,6,7,9 | **5,8,9** |
| Vowel | 4,5,6,7 | 4,5,6,7,9 | 3,4,5,6,8 | **2,3,5** |
| WBC | 2,4,5,6,8 | 3,4,5,6,7 | 4,5,6,7,8 | **3,5** |
| Wine | 4,6,7,8,10 | 5,6,7,9 | 3,4,5,6,7,8,9 | **2,5,8** |
| Heart | 4,5,6,7,9 | 3,5,6,7,9 | 5,6,7,8,9,10 | **3,4,6,13** |
| Segment | 6,7,9,10,11,12 | 3,8,9,11,12 | 7,9,10,11,12 | **14,15,17** |
| Sonar | 26,28,32,34,37 | 27,30,32,35,36 | 20,26,28,32,34,36 | 18,5,10,44,22,26,38,34,37,50 |
| Two norm | 9,10,11,12,14,16 | 7,9,10,11,13,14 | 11,12,13,14,15,17 | 2,7,10,14,18,19 |

The results of classification based on the method proposed in this study are shown in Table 4. As Fig. 3 indicates, ISPSO has greater quality than the other methods. By using the nearest neighbor (1-NN) for evaluating each single particle, the accuracy of the proposed method was estimated. As the information in Table 4 shows, method proposed in this study had higher classification accuracy than other methods of feature selection. As an instance, the mean accuracy and standard deviation for 1-NN classifier was reported to be 97.30 and 3.08, respectively, but mean accuracy and standard deviation for PSO, ACO, SA, and GA were reported to be 82.23 (3.71), 66.9 (8.16), 76.83 (3.48) and 83.28 (5.36), respectively. Furthermore, when no feature selection method was applied, mean accuracy level for the use of original features was 81.09. Moreover, the best obtained results are indicated in boldface.

**Table 4**

Shows mean accuracy in classification and SD for ISPSO-GLOBAL, PSO, ACO, SA, and GA methods of selecting features during 20 separate runs by the use of 1-NN classifier. In the dataset, "Without FS" indicates the time all features are included in the experiment. Moreover, the best obtained results are indicated in boldface.

| Data sets | Without FS | GA | SA | ACO | PSO | HPSO-LS | IPSO-GLOBAL |
|---|---|---|---|---|---|---|---|
| Glass | 67.09 | 64.51 ± 8.86 | 62.64 ± 4.7 | 70.77 ± 6.54 | 70.31 ± 3.68 | 74.91 ±3.63 | **76.00 ±12.9** |
| Vowel | 97.60 | 84.109 ± 12.47 | 84.03 ± 12.53 | 74.857 ± 14.06 | 95.872 ± 0.26 | 99.87 ± 0.20 | **100 ± 0.00** |
| WBC | 96.12 | 96.361 ± 2.42 | 96.88 ± 1.03 | 95.833 ± 1.31 | 97.69 ± 0.58 | **98.27 ± 0.4** | 96.93 ± 4.55 |
| Wine | 73.02 | 92 ± 8.02 | 89.18 ± 3.86 | 92.236 + 2.724 | 94.24 ± 4.18 | 97.17 ± 1.39 | **97.47 ± 3.10** |
| Heart | 95.46 | 82.388 ± 0.1218 | 85.81 ± 0.130 | 89.87 ± 0.1328 | 84.42 ± 0.166 | 78.84 ± 2.07 | **90. 70 ± 1.10** |
| Segment | 96.03 | 83.11 ± 5.08 | 81.181 ± 9.74 | 92.17 ± 0.74 | 90.775 ± 0.93 | 92.348 ± 0.669 | **95.35 ± 1. 83** |
| Two norm | 93.24 | 86.35 ± 3.9 | 85.13 ± 3.09 | 74.9 ± 6.88 | 89.92 ± 2.82 | 94.25 ± 0.59 | **96.64± 5.06** |
| Sonar | 81.06 | 83.28 ± 5.36 | 76.83 ± 3.48 | 66.9 ± 8.16 | 82.233 ± 3.71 | 87.23 ± 3.40 | **97.30 ± 3.08** |
| Arrhythmia | 49.62 | 49.69 ± 5.33 | 48.99 ± 3.61 | 47.99 ± 5.90 | 51.65 ± 4.51 | 53 ± 2.28 | **73.86±36.01** |
| LSVT | 60.52 | 71.71 ± 4.93 | 71.45 ± 3.01 | 68.59 ± 6.26 | 73.21 ± 4.89 | 77.06 ± 3.62 | **79.92±15.42** |
| Lymphoma | 61.53 | 78.02±4.43 | 76.83±6.65 | 69.63±5.43 | 80.65±4.76 | **82.85 ± 5.48** | 80.73±5.54 |

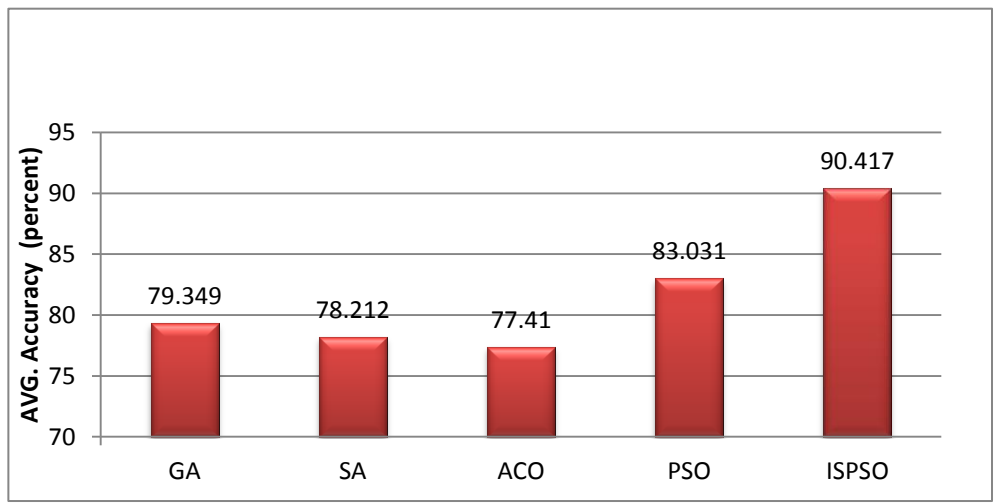

Fig. 3. Average classification accuracies for different features subset sizes of GA, SA, ACO, PSO and ISPSO-GLOBAL feature selection methods.

ISPSO-GLOBAL method was also compared with a new method [57]. Table 5 shows the results of eleven data sets utilized in this paper. The results in Table 5 show that, the highest accuracy rate was obtained by using the method proposed in this study. As an example, the results of this study yielded that the mean accuracy and standard deviation of 1-NN classifier in comparison with the Zoo dataset was 98.13 and .045, respectively. However, the mean accuracy (and standard deviation) for DFS, r2PSO, r3PSO, and rPSO-1hc was reported to be 96.87 (.025), 97.7 (.031), 96.5 (.03), and 97.75 (.01), respectively. Furthermore, with no application of feature selection methods, the mean accuracy rate for the use of original features was reportedly 93.02. The mean accuracy in classification and standard deviation of ISPSO-GLOBAL, PSO, ACO, GA, and SA methods of feature selection run with 1-NN classifier are presented in Table 5.

**Table 5**
Features were identified by various feature selection methods over 20 independent runs using 1-NN classifier.
Note that CA refers to the average classification accuracy.

| Datasets | Measure | Without FS | DFS | r2PSO | r3PSO | r2PSO-lhc | **IPSO-GLOBAL** |
|---|---|---|---|---|---|---|---|
| WBCO | Fea.NO | 9 (0) | 2.05 (0.6048) | 1.2 (0.41) | 2.75 (0.4443) | 1.45 (0.51) | 2(0) |
| | CA | 0.96 (0.004) | 0.9337(0.019) | 0.9340 (0.012) | 0.8741 (0.046) | 0.939 (0.01) | **0.9692(0.1393)** |
| Glass | Fea.NO | 9 | 3.4 (0.695) | 3.4 (0.502) | 3.35 (0.5871) | 3.5 (0.6882) | 3(0) |
| | CA | 0.671 (0.05) | 0.7157 (0.044) | 0.7186 (0.043) | 0.7104 (0.05) | 0.786 (0.03) | 0.7600(0.1393) |
| Wine | Fea.NO | 13 (0) | 2.65 (0.67) | 2.7 (0.6569) | 2.55 (0.6048) | 2.6 (0.5982) | 3(0) |
| | CA | 0.732 (0.03) | 0.9514 (0.021) | 0.9563 (0.019) | 0.95 (0.017) | 0.9563(0.02) | **0.9747(3.10)** |
| Zoo | Fea.NO | 16 (0) | 4.6 (1.04) | 3.9 (0.8335) | 4.5 (0.6882) | 4.75 (1.019) | 5(0) |
| | CA | 0.932 (0.06) | 0.9687 (0.025) | 0.9775 (0.031) | 0.965 (0.0347) | 0.9775(0.01) | **0.9813 (0.0450)** |
| Vehicle | Fea.NO | 19 | 5.6 (1.142) | 4.8 (0.6958) | 4.8 (0.6958) | 5.15 (0.875) | 5(0) |
| | CA | 0.85 (0.217) | 0.7336 (0.021) | 0.7270 (0.013) | 0.7302 (0.02) | 0.7303 (0.01) | **0.8905(0.3279)** |
| Protein | Fea.NO | 20 (0) | 7.75 (1.2926) | 7.25 (1.5517) | 7.95 (1.468) | 7.95 (1.6375) | 6(0) |
| | CA | 0.691(0.068) | 0.7880 (0.045) | 0.8173 (0.042) | 0.8348 (0.047) | 0.8195 (0.04) | **0.8854** |
| Segment | Fea.NO | 20 (0) | 3.3 (0.4701) | 3 (0) | 3 (0) | 3 (0) | 3 (0) |
| | CA | 0.96 (0.007) | 0.9674 (0.006) | 0.9665 (0.004) | 0.9693 (0.005) | 0.966 (0.004) | **0.9745 (0.0321)** |
| Wdbc | Fea.NO | 30 (0) | 4.95 (0.9986) | 2.55 (0.6863) | 2.9 (0.7181) | 2.6 (0.6805) | 2(0) |
| | CA | 0.916 (0.01) | 0.9342 (0.008) | 0.9418 (0.01) | 0.9432 (0.008) | 0.9381 (0.01) | **0.9685 (0.0321)** |
| Soybean | Fea.NO | 35(0) | 4.75 (1.5174) | 2 (0) | 1.9 (0.307) | 2 (0) | 4(0) |
| | CA | 0.984 (0.03) | 1 (0) | 1 (0) | 1 (0) | 1 (0) | 1(0) |
| Lung cancer | Fea.NO | 56 (0) | 19.4 (2.5214) | 9.35 (2.996) | 7.9 (2.1496) | 13.1 (2.3373) | **7(0)** |
| | CA | 0.461 (0.13) | 0.9269 (0.046) | **0.9846 (0.031)** | **0.9846 (0.031)** | 0.95 (0.0516) | 0.9666(0.4516) |
| Sonar | Fea.NO | 60 (0) | 18.75 (2.6532) | 12.2 (2.1908) | 11.85 (1.7252) | 14.4 (2.0365) | 11(0) |
| | CA | 0.816 (0.02) | 0.9265 (0.023) | 0.9668 (0.018) | 0.9674 (0.017) | 0.951 (0.02) | **0.9790(0.0813)** |

Fig. 4 shows average classification accuracies for different features subset sizes of ISPSO-GLOBAL, PSO, ACO, SA, GA and HPSO-STS, SPSO-QR, PSO (4-2) and CBPSO feature selection methods. Moreover, mean classification accuracy and SD for PSO (4-2), CBPSO, SPSO-QR, HPSO-STS, and HGAFS methods of feature selection by the use of 1-NN classifier are shown in Table 6.

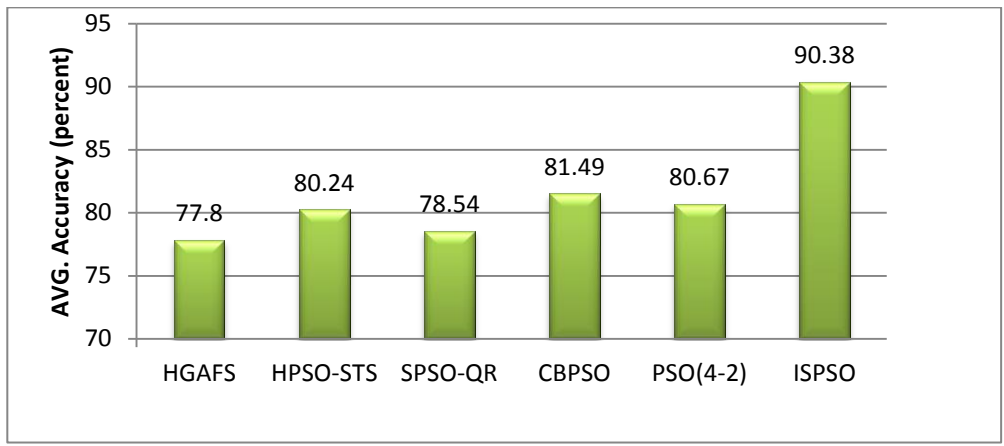

Fig. 4. Average classification accuracies for different features subset sizes of HGAFS, HPSO-STS, SPSO-QR, CBPSO, ISPSO-GLOBAL and PSO (4-2) feature selection methods.

**Table 6**
Average classification accuracy (percent and standard deviation of ISPSO-GLOBAL, PSO(4-2), CBPSO, SPSO-QR, HPSO-STS and HGAFS feature selection methods over 20 independent runs using 1-NN classifier. In each data set, the best method is marked by the boldface and underlined and the second best result is also boldfaced.

| Data sets | HGAFS | HPSO-STS | SPSO-QR | CBPSO | PSO(4-2) | HPSO-LS | ISPSO-GLOBAL |
|---|---|---|---|---|---|---|---|
| Glass | 65.16 ± 2.30 | 67.57 ± 6.06 | 70.75 ± 7.25 | 68.74 ± 6.54 | 73.21 ± 6 | 74.91 ±3.63 | **76.00 ± 12.95** |
| Vowel | 87.34 ± 2.43 | 92.96 ± 13.09 | 89.68 ± 11.90 | 98.47 ± 1.61 | 96.01 ± 5.37 | **99.87 ±0.20** | **100 ± 0.00** |
| WBC | 96.78 ± 0.74 | 95.73 ± 3.70 | 96.38 ± 1.43 | **97.93 ± 0.30** | 97.04 ± 1.33 | **98.27 ±0.4** | 96.93 ± 4.55 |
| Wine | 90.68 ± 7.80 | 96.72 ± 0.97 | 91.59 ± 7.28 | 95.17 ± 4.04 | 93.27 ± 3.27 | **97.17 ±1.39** | **97.47 ± 3.10** |
| Heart | 68.11 ± 7.72 | 73.65 ± 7.05 | 74.22 ± 7.12 | 76 ± 5.60 | 73.78 ± 4.13 | **78.84 ±2.07** | **90. 70 ± 1.10** |
| Segment | 88.49 ± 1.11 | 88.19 ± 7.02 | 81.33 ± 9.08 | 90.03 ± 1.36 | 85.42 ± 4.81 | **92.348 ±0.669** | **95.35 ± 1.83** |
| Two norm | 87.14 ± 4.48 | 91.43 ± 1.97 | 85.92 ± 3.23 | 89.59 ± 2.53 | 89.29 ± 2.67 | **94.25 ±0.59** | **96.64 ± 5.06** |
| Sonar | 79.20 ± 4.17 | 82.65 ± 4.29 | 80.28 ± 5.02 | 84.30 ± 3.38 | 83.24 ± 3.75 | **87.23 ±3.40** | **97.30 ± 3.08** |
| Arrhythmia | 47.93 ± 2.83 | 50.37 ± 1.04 | 50.05 ± 0.722 | 48.96 ± 2.99 | 48.14 ± 3.87 | **53.00 ±2.28** | **73.86±36.01** |
| LSVT | 68.05 ± 7.16 | 63.15 ± 3.72 | 65.20 ± 2.87 | 65.78 ± 1.99 | 67.36 ± 7.11 | **77.06 ±3.62** | **79.92 ± 15.42** |
| Average | 77.8 ± 4.074 | 80.24 ± 4.891 | 78.54 ± 5.59 | 81.49 ± 3.034 | 80.67 ± 4.231 | **84.20 ±1.54** | **90.38 ± 10.074** |

Table 7 shows the possible average classification accuracy of ISPSO-GLOBAL and those of filter-based methods including information gain, Fisher-score, term variance and mRMR over 20 independent runs using 1-NN. According to Table 8 the p-value results of T-test between the classification performance achieved by the ISPSO-GLOBAL over 20 independent runs and those of CBPSO, SPSO-QR, HPSO-STS and HGAFS feature selection methods. We compare the performance of the proposed method with four recently published feature selection methods and the results are shown in Tables 7 and 8. The results also show that the proposed method outperformed the others in terms of the classification accuracy.

**Table 7**

Average classification accuracy (percent) of ISPSO-GLOBAL and those of filter based methods including information gain (IG), fisher score (F-Score), term variance (TV) and mRMR over 20 independent runs using 1-NN classifier. Best results are boldfaced.

| Data sets | HGAFS | HPSO-STS | SPSO-QR | CBPSO | PSO(4-2) | ISPSO-GLOBAL |
|---|---|---|---|---|---|---|
| Glass | 65.16 ± 2.30 | 67.57 ± 6.06 | 70.75 ± 7.25 | 68.74 ± 6.54 | **73.21 ± 6** | **76.00 ± 12.95** |
| Vowel | 87.34 ± 2.43 | 92.96 ± 13.09 | 89.68 ± 11.90 | **98.47 ± 1.61** | 96.01 ± 5.37 | **100 ± 0.00** |
| WBC | 96.78 ± 0.74 | 95.73 ± 3.70 | 96.38 ± 1.43 | **97.93 ± 0.30** | 97.04 ± 1.33 | 96.93 ± 4.55 |
| Wine | 90.68 ± 7.80 | **96.72 ± 0.97** | 91.59 ± 7.28 | 95.17 ± 4.04 | 93.27 ± 3.27 | **97.47 ± 3.10** |
| Heart | 68.11 ± 7.72 | 73.65 ± 7.05 | 74.22 ± 7.12 | **76 ± 5.60** | 73.78 ± 4.13 | **90. 70 ± 1.10** |
| Segment | 88.49 ± 1.11 | 88.19 ± 7.02 | 81.33 ± 9.08 | **90.03 ± 1.36** | 85.42 ± 4.81 | **95.35 ± 1.83** |
| Two norm | 87.14 ± 4.48 | **91.43 ± 1.97** | 85.92 ± 3.23 | 89.59 ± 2.53 | 89.29 ± 2.67 | **96.64 ± 5.06** |
| Sonar | 79.20 ± 4.17 | 82.65 ± 4.29 | 80.28 ± 5.02 | **84.30 ± 3.38** | 83.24 ± 3.75 | **97.30 ± 3.08** |
| Arrhythmia | 47.93 ± 2.83 | **50.37 ± 1.04** | 50.05 ± 0.722 | 48.96 ± 2.99 | 48.14 ± 3.87 | **73.86±36.01** |
| LSVT | 68.05 ± 7.16 | 63.15 ± 3.72 | 65.20 ± 2.87 | 65.78 ± 1.99 | **67.36 ± 7.11** | **79.92 ± 15.42** |
| Average | 77.8 ± 4.074 | 80.24 ± 4.891 | 78.54 ± 5.59 | **81.49 ± 3.034** | 80.67 ± 4.231 | **90.38 ± 10.074** |

**Table 8**

The p-value results of statistical T-test. The results lower that 0.005 are bolded.

| Data sets | HGAFS | HPSO-STS | SPSO-QR | CBPSO | PSO(4-2) | ISPSO-GLOBAL |
|---|---|---|---|---|---|---|
| Glass | 65.16 ± 2.30 | 67.57 ± 6.06 | 70.75 ± 7.25 | 68.74 ± 6.54 | **73.21 ± 6** | **76.00 ± 12.95** |
| Vowel | 87.34 ± 2.43 | 92.96 ± 13.09 | 89.68 ± 11.90 | **98.47 ± 1.61** | 96.01 ± 5.37 | **100 ± 0.00** |
| WBC | 96.78 ± 0.74 | 95.73 ± 3.70 | 96.38 ± 1.43 | **97.93 ± 0.30** | 97.04 ± 1.33 | 96.93 ± 4.55 |
| Wine | 90.68 ± 7.80 | **96.72 ± 0.97** | 91.59 ± 7.28 | 95.17 ± 4.04 | 93.27 ± 3.27 | **97.47 ± 3.10** |
| Heart | 68.11 ± 7.72 | 73.65 ± 7.05 | 74.22 ± 7.12 | **76 ± 5.60** | 73.78 ± 4.13 | **90. 70 ± 1.10** |
| Segment | 88.49 ± 1.11 | 88.19 ± 7.02 | 81.33 ± 9.08 | **90.03 ± 1.36** | 85.42 ± 4.81 | **95.35 ± 1.83** |
| Two norm | 87.14 ± 4.48 | **91.43 ± 1.97** | 85.92 ± 3.23 | 89.59 ± 2.53 | 89.29 ± 2.67 | **96.64 ± 5.06** |
| Sonar | 79.20 ± 4.17 | 82.65 ± 4.29 | 80.28 ± 5.02 | **84.30 ± 3.38** | 83.24 ± 3.75 | **97.30 ± 3.08** |
| Arrhythmia | 47.93 ± 2.83 | **50.37 ± 1.04** | 50.05 ± 0.722 | 48.96 ± 2.99 | 48.14 ± 3.87 | **73.86±36.01** |
| LSVT | 68.05 ± 7.16 | 63.15 ± 3.72 | 65.20 ± 2.87 | 65.78 ± 1.99 | **67.36 ± 7.11** | **79.92 ± 15.42** |
| Average | 77.8 ± 4.074 | 80.24 ± 4.891 | 78.54 ± 5.59 | **81.49 ± 3.034** | 80.67 ± 4.231 | **90.38 ± 10.074** |

## 5. Conclusion and Future Directions

In this study, a feature selection method called ISPSO-GLOBAL is proposed. The proposed method is able to select a small number of most prominent features ISPSO-GLOBAL method plays an important role in the classification task to reduce the computational cost, simplify the learning model and improve the general abilities of classifiers. The proposed method develops a novel initialization strategy motivated by chaos theory. Also a novel Pbest and Gbest updating mechanisms are proposed to consider both the number of feature and fitness particles to overcome the limitation of the traditional updating mechanism of PSO. We postulate the probability of a given state as a function of the vote's features; it must satisfy the relationship between good features in next generation. ISPSO-GLOBAL achieved significantly better classification performance than using all features selection methods. In other words, we have a tendency to in classification those datasets with a smaller number of features and shorter procedure time. In the future, we are able to use the $v$ and $r$ to different feature choice strategies or combinatorial optimization strategies in different domains.

However, in future, we aim to tryout the $r$ and $v$ on large feature subsets. Moreover, we tend to the improved searching ability for the feature selection task.